\documentclass[review]{elsarticle}
 
\usepackage[T1]{fontenc}
\usepackage{libertine}
\usepackage[scaled=0.85]{beramono}

\usepackage{paralist}

\newcommand{\bi}{\begin{compactitem}}
\newcommand{\ei}{\end{compactitem}}
\usepackage{wrapfig}
\usepackage{dblfloatfix}
\usepackage{subcaption}
\usepackage{lipsum}
\usepackage{tabularx}
\usepackage{multicol}
\usepackage{amsmath}
\usepackage{amsfonts}
\usepackage{amsthm}
\usepackage[ruled,linesnumbered]{algorithm2e}
\usepackage{bm, bbm, dsfont}
\usepackage{colortbl}
\usepackage{framed}
\usepackage{changepage}
\usepackage{multirow}
\usepackage[shortlabels]{enumitem}
\usepackage{booktabs}
\usepackage{tikz}
\usepackage[many,breakable]{tcolorbox}
\usepackage{url}

\usepackage{enumitem}
\newlist{todolist}{itemize}{2}
\setlist[todolist]{label=$\square$}
\usepackage{pifont}

\tcbuselibrary{skins}
\newtheorem{definition}{Definition}
\newtcolorbox{blockquote}{colback=blue!5!white,
                          boxrule=0.4pt,
                          colframe=gray!60!black,
                          fonttitle=\bfseries,
                          drop fuzzy shadow,
                          outer arc=3pt}

\DeclareMathOperator*{\argmax}{arg\,max}
\DeclareMathOperator*{\argmin}{arg\,min}

\makeatletter
 {\par\unskip\endMakeFramed}
\makeatother

\definecolor{formalshade}{rgb}{0.93,0.93,0.93}

\definecolor{darkblue}{rgb}{0.2, 0.2, 0.2}

\newenvironment{formal}{%
  \def\FrameCommand{%
    \hspace{1pt}%
    {\color{darkblue}\vrule width 2pt}%
    {\color{formalshade}\vrule width 4pt}%
    \colorbox{formalshade}%
  }%
  \MakeFramed{\advance\hsize-\width\FrameRestore}%
  \noindent\hspace{-1pt}
  \begin{adjustwidth}{}{7pt}%
  \vspace{2pt}\vspace{2pt}%
}
{%
  \vspace{3pt}\end{adjustwidth}\endMakeFramed%
}

\newtcolorbox{answer}{colback=gray!10!white,boxrule=0.4pt,colframe=gray!60!black,fonttitle=\bfseries}
\newcolumntype{L}{>{\raggedright\arraybackslash}X}
\newtcolorbox{todo}{colback=yellow!60,colframe=red!30}

\definecolor{fill}{rgb}{0.85, 0.85, 0.85}
\newcommand{\gray}[1]{\cellcolor{fill}\textcolor{black}{#1}}

\AtBeginDocument{%
  \providecommand\BibTeX{{%
    \normalfont B\kern-0.5em{\scshape i\kern-0.25em b}\kern-0.8em\TeX}}}

\newcommand{\HPO}{hyper-parameter optimization~}
\newcommand{\ITT}{\tt{DEEPLY}}
\newcommand{\IT}{{\ITT}~}

\journal{Expert Systems with Applications}

\begin{document}
\begin{frontmatter}









\title{An Expert System for Redesigning Software for Cloud Applications }

\author[ncsu]{Rahul Yedida \corref{cor1}}
\ead{ryedida@ncsu.edu}

\author[ibm]{Rahul Krishna}
\ead{rkrsn@ibm.com}

\author[ibm]{Anup Kalia}
\ead{anup.kalia@ibm.com}

\author[ncsu]{Tim Menzies}
\ead{timm@ieee.org}

\author[ibm]{Jin Xiao}
\ead{jinoaix@us.ibm.com}

\author[ibm]{Maja Vukovic}
\ead{maja@us.ibm.com}

\cortext[cor1]{Corresponding author.}
\address[ncsu]{Department of Computer Science, North Carolina State University, USA}
\address[ibm]{IBM Research, USA}

\begin{abstract}
 Cloud-based
software has  many advantages.  
When services are divided into many  independent components, they are easier to update. Also, during
 peak demand, it is  easier to scale cloud services (just hire more CPUs).  Hence,  many organizations are
partitioning their 
     monolithic  enterprise applications into  
     cloud-based microservices.

   Recently there has been much  work using
   machine learning  to simplify this  partitioning
   task. Despite much research, 
     no single partitioning method can be recommended as generally useful.
     More specifically,   those
     prior solutions 
     are ``brittle''; i.e. if they  work well for one kind of goal in one
     dataset,  then they  can be  sub-optimal
      if  applied
     to many datasets and multiple  goals.    
    
In order to find a generally
useful partitioning method,
we propose  {\IT}.
This new algorithm extends  the CO-GCN deep learning partition
  generator with (a)~a novel loss function 
  and (b)~some hyper-parameter optimization.
    As  shown by our   experiments, {\IT}  generally outperforms prior work (including CO-GCN, and  others)
   across multiple datasets and goals. To the best of
our knowledge, this is the first report in SE  of such stable hyper-parameter optimization.
    
 To enable the reuse of this research,  {\IT} is available on-line at \url{https://bit.ly/2WhfFlB}. 
\end{abstract}



%

\end{frontmatter}

\section{Introduction}
\label{sec:introduction}

As more and more enterprises move to the cloud, new tools are needed. For example, IBM helps clients with millions of lines of code each year in this refactoring process. In one such effort, IBM worked with a Fortune 100 company to recommend partitions for a system with over one million lines of code. These partitions were manually inspected by subject matter experts and verified within weeks (as opposed to a year of manual effort)\footnote{\url{http://tiny.cc/mono2micro}}. Unfortunately, the tool support needed for this process is still in its infancy. For example, consider the problem of how to divide up old software for the cloud (i.e., into microservices). Informally, we need to encourage cohesion and minimize coupling. However, AI-based tools for doing this require more precise definitions of cohesion and coupling. In this paper, we show six state-of-the-art tools focused on that exact problem, which internally have several internal hyper-parameter choices. The challenge with these tools is to tame that large hyper-parameter space.

Data mining is a powerful tool but, like any other software
system~(\cite{xu15},
analysts are often puzzled by all the options for
the control settings. For example,  consider the task
of  converting  monolithic enterprise
software into   cloud microservices.  For the task, 
it is a common practice to apply some  clustering algorithm to
decide how to break up the code into $k$ smaller microservices.
A   common question asked by programmers is ``what is a good
value for $k$"?
More generally, across all the learners used for microservice partitioning, currently there is little support for selecting  
appropriate control settings (\cite{desai2021graph,yedida2021lessons}.

Tools that can automatically learn settings for data miners
are called {\em hyper-parameter optimizers} (HPO). These tools can learn
(e.g.)
good $k$ values while also optimizing for other goals
including cluster coherence (which should
be maximized) and coupling (which should be minimized).  But
HPO suffers from 
  {\em hyper-parameter brittleness}.
  For example, 
\citet{tu2018one}  reported that if an optimizer works well for one kind of goal in one data set, they can be sub-optimal if applied to multiple datasets and goals.
In the case of redesigning  software monoliths
as cloud microservices,
\citet{yedida2021lessons}
recently reported that different HPO tools perform best for different sets
of goals being explored on different datasets.
To say that another way, based on past results, no specific prior partitioning method can be recommended as generally useful.
      This we consider this as a significant problem. As designs get more
      complex, partitioning methods become very slow~(\cite{yedida2021lessons}.
      For example,  at the time of this writing, we are  
      running our  algorithms for an industrial client.
      That  process has taken  282 CPU hours for 1 application.  Hence it is less-than-ideal to ask engineers to hunt through the output of  multiple
      partitioning algorithms, looking
      for results that work best for their
      particular domain. This especially true  when each of
      those algorithms runs very slowly. Instead,
      we should be able to offer them one partitioning method that is generally most useful across a wide range of problems.

To find a generally useful partitioning methods, this paper seeks HPO tools that perform best across multiple  datasets and 
goals
(and prior work~(\cite{kalia2021mono2micro,jin2019service,mazlami2017extraction,mitchell2006automatic,desai2021graph} tended to explore  just one or two partitioning methods). Thus, we propose {\IT}, which is a novel
 combination
of optimization (using Bergstra's  hyperopt tool~(\cite{bergstra2013hyperopt}) and a loss function. 
As shown by ur results, {\IT}
generally works well across multiple goals and data sets.   

To understand the benefits of {\IT}, we investigate  five research questions.

{\bf RQ1:}  {\em How prevalent is hyper-parameter brittleness in automated microservice partitioning?}
We verify \citet{yedida2021lessons}'s results, who showed that hyper-parameter optimizers are ``brittle", i.e., they work well for a few metrics and datasets, but are not useful across
multiple goals and data sets. The verification is crucial to set the motivation for our contribution.


 {\bf RQ2:} {\em  Is hyper-parameter optimization enough to curb the aforementioned optimizer brittleness?}
 Here we will show that   standard  hyper-parameter optimization methods are insufficient for
 solving brittleness.

{\bf RQ3:} Since hyper-parameter optimization methods are not enough,  {\em How else might we fix the aforementioned brittleness problem?} To that purpose, we propose {\IT}, a novel
combination of   hyper-parameter optimizers with a new loss function.

{\bf RQ4:}  {\em Does {\IT}   generate ``dust'' 
 (where  all functionality is loaded into one partition) or ``boulders'' (where all partitions contain only one class each)?}  Here, we show that {\IT} avoids two anti-patterns. Specifically, {\IT} does not create   ``boulders'' or ``dust''.

The rest of this paper is structured as follows. We provide a detailed background on the problem and the various attempts at solving it in \S \ref{sec:background}. We then formalize the problem and discuss our method in \S \ref{sec:partitioning}. In \S \ref{sec:setup}, we discuss our experimental setup and evaluation system. We present our results in \S \ref{sec:results}. Next, we show how the crux of our approach extends beyond this problem in \S \ref{sec:broader}. We discuss threats to the validity of our study in \S \ref{sec:threats}. Finally, we conclude in \S \ref{sec:conclusion}. 

Before we begin,
just to say the obvious,
when we say that  {\HPO} is becomes more
generally useful method  with {\IT},   we mean ``generally useful across the data sets and metrics \underline{\bf explored thus far}''.
It is an open question, worthy of    future research,
to test if our methods apply to  other   datasets and goals.

\section{ Designing for the Cloud}
\label{sec:background}

To fully exploit the cloud,
systems have to be rewritten as  ``microservices''
comprising  multiple independent, loosely coupled pieces that can scale independently.  Microservice architectures have many advantages (\cite{al2018comparative,wolff2016microservices} such as technology heterogeneity, resilience (i.e., if one service fails, it does not bring down the entire application), and ease of deployment. Therefore, it is of significant business interest to port applications to the cloud under the microservice architecture. For example, Netflix states that the elasticity of the cloud and increased service availability are two of the primary reasons it switched to a microservice architecture\footnote{\url{https://about.netflix.com/en/news/completing-the-netflix-cloud-migration}}. They use the cloud for distributed databases, big data analytics, and business logic. Further, they mention that moving to the cloud reduced their averaged cost.
.  

\citet{daya2016microservices} report  on  the long queue of applications
waiting to be ported to the cloud.  Once ported to cloud-based microservices,   code under each microservice can be independently enhanced and scaled, providing agility and improved speed of delivery.  
But converting 
traditional enterprise software to microservices is problematic.
Such traditional code is
{\em   monolithic}, i.e., all the components are tightly coupled, and the system is designed as one large entity.
Thus, some
{\em application refactoring} process is required
to convert the monolithic code into to  a set of small code
clusters.

\begin{table*}[!t]
    \centering
    \caption{
    \citet{yedida2021lessons} report five widely used metrics to   assess the quality of microservice partitions. In this table,
     [-] denotes metrics where {\em less} is better and
       [+] denotes metrics where {\em more} is better. For details on how these metrics are defined, see \S\ref{sec:setup:metrics}.}
    \label{tab:metrics} { \footnotesize
    \begin{tabularx}{\textwidth}{LlLL}
        \toprule
        \textbf{Metric} & \textbf{Name} & \textbf{Description} & \textbf{Goal} \\
        \midrule
        BCS [-] (\cite{kalia2020mono2micro} & Business context sensitivity & Mean entropy of business use cases per partition & If \textbf{minimized} then more business cases handled locally \\
        ICP [-] (\cite{kalia2020mono2micro} & Inter-partition call percentage & Percentage of runtime calls across partitions & If \textbf{minimized} then less traffic between clusters \\
        SM [+] (\cite{jin2019service} & Structural modularity & A combination (see \S\ref{sec:setup:metrics}) of cohesion and coupling defined by \citet{jin2019service} & If \textbf{maximized} then more self-contained clusters with fewer connections between them \\
        MQ [+] (\cite{mitchell2006automatic} & Modular quality & A ratio involving cohesion and coupling, defined by \citet{mitchell2006automatic} & If \textbf{maximized} then more processing is local to a cluster \\
        IFN [-] (\cite{mitchell2006automatic} & Interface number & Number of interfaces needed in the microservice architecture & If \textbf{minimized} then fewer calls between clusters \\ 
        NED [-] (\cite{kalia2020mono2micro} & Non-extreme distribution & Number of partitions with non-extreme values & If \textbf{minimized}, fewer ``boulders" (large monolithic partitions) or ``dust'' (partitions with only one class)\\
        \bottomrule
    \end{tabularx} }
\end{table*}
 
\begin{table}[!t]
    \centering
    \caption{Hyper-parameter choices  studied in this paper.
    All the methods take runtime traces
(or use cases) as their input. These methods
return suggestions on how to build partitions from
all the  classes seen
in the traces. For further details on these
partitioning algorithms, see 
\S\ref{sec:partitioning}.}
    \label{tab:parameters} { \footnotesize
    \begin{tabularx}{\linewidth}{lLl}
        \toprule
        \textbf{Algorithm} & \textbf{Hyper-parameter} & \textbf{(min, default, max)}  \\
        \midrule
        Mono2Micro (\cite{kalia2020mono2micro, kalia2021mono2micro}) & Number of clusters & (2, 5, 10)  \\
        \midrule
        \multirow{5}{*}{FoSCI (\cite{jin2019service})} & Number of clusters & (5, 5, 13) \\
        & Number of iterations to run NSGA-II & (1, 3, 6) \\
        & Population size for NSGA-II & (30, 100, 200) \\
        & Parent size to use in NSGA-II & (10, 30, 50) \\
        \midrule
        \multirow{2}{*}{MEM (\cite{mazlami2017extraction})} & Maximum partition size & (17, 17, 17) \\
        & Number of partitions & (2, 5, 13) \\
        \midrule
        \multirow{3}{*}{Bunch (\cite{mitchell2006automatic})} & Number of partitions & (3, 5, 13) \\
        & Initial population size for hill climbing & (2, 10, 50) \\
        & Number of neighbors to consider in hill-climbing iterations & (2, 5, 10) \\
        \midrule
        \multirow{4}{*}{CoGCN (\cite{desai2021graph})} & Number of clusters & (2, 5, 13) \\
        & Loss function coefficients $\alpha_1, \alpha_2, \alpha_3$ & (0, 0.1, 1) \\
        & Number of hidden units in each layer, $h_1, h_2$ & (4, 32, 64) \\
        & Dropout rate & (0, 0.2, 1) \\
        \bottomrule
    \end{tabularx} }
\end{table}

When done manually, such  refactoring is  expensive, time-consuming, and error-prone.
Hence, there is much current interest in automatically refactoring traditional
systems into cloud services. 

 Some of the authors of this paper are part of the IBM ``Mono2Micro'' that helps client redesign their systems for the cloud. Some of that analysis is manual since it replies on extensive domain knowledge. That said, increasingly, there is automation applied to this task. For example, given a set of test cases or use cases, it is reasonable to ask ``how does this knowledge of frequently use tests or use cases inform our microservices design?''. For that purpose, some AI clustering can be used. 
 
 But when we try to use AI tools for this task,
 we often encounter the same problem. 
 Specifically, there are so
 many   algorithms and partitioning goals:
\bi
\item
  Table~\ref{tab:metrics}
   lists the various coupling and cohesion goals
 used by prior work on partition generation. Note that
 in column one, any goal marked with ``[-]'' should
 be {\em minimized} and all other goals should 
 be {\em maximized}.
 \item
See also Table~\ref{tab:parameters} which lists some of the
 partitioning tools, as well as their control parameters.   In the table,  the core
idea is to understand what parts
 of the code call each other
 (e.g., by reflecting on test cases
 or use cases), then cluster those
 parts of the code into separate microservices.
 These algorithms assess the value of different partitions
 using scores generated from the Table~\ref{tab:metrics} metrics.
\ei
In such a rich   space or options, different state-of-the-art AI approaches may generate different recommendations.
\citet{yedida2021lessons} concluded that looking at prior work, we seem to have a situation
where   analysts might
run several  partitioning algorithms to find
an algorithm that performs the best based on their business requirements. For example, analysts might prefer Mono2Micro (\cite{kalia2020mono2micro, kalia2021mono2micro}) over others considering the clean separation of business use cases where as they might prefer CoGCN (\cite{desai2021graph}) and FoSCI (\cite{jin2019service}) for low coupling.
As stated in the introduction, this is hardly ideal since these
partitioning algorithms can be slow to execute. Accordingly,
this paper seeks optimization methods that support partitioning
for a wide range of data sets and goals.


\section{Algorithms for Microservice Partitioning}
\label{sec:partitioning}

Informally, partitioning algorithms   take domain knowledge and
propose partitions containing classes that often connect to each other.
For example, given a set of use cases or traces of test case execution:
\bi
\item Identity the  entities (such as classes) used in   parts of the use cases/test cases;
\item Aggregate the frequently connecting entities;
\item Separate the entities that rarely connect.
\ei
This can be formalized as follows. Consider classes\footnote{This definition trivially extends to languages without classes; translation units such as functions could be used instead, for example.} in an application $A$ as $\mathcal{C}^A$ such that $\mathcal{C}^A = \{ c_1^A, c_2^A, \ldots, c_k^A \}$, where $c_i^A$ represents an individual class. With this, we define a partition as follows:

\begin{definition}
A partition $\mathcal{P}^A$ on $\mathcal{C}^A$ is a set of subsets
\newline $\{ P_1^A, P_2^A, \ldots, P_n^A \}$ such that
\begin{itemize}
    \item $\bigcup\limits_{i=1}^n P_i^A = \mathcal{C}^A$, i.e., all classes are assigned to a partition.
    \item $P_i^A \neq \phi~\forall i=1,\ldots,n$, i.e., there are no empty partitions.
    \item $P_i^A \cap P_j^A = \phi~\forall i \neq j$, i.e., each partition is unique.
\end{itemize}
\end{definition}

\begin{definition}
A partitioning algorithm $f$ is a function that induces a partition $\mathcal {P}_A$ on an application with classes $\mathcal{C}^A$.
\end{definition}

The goal of a microservice candidate identification algorithm is to identify a function $f$ that jointly maximizes a set of metrics $m_1, m_2, \ldots, m_p$ that quantify the quality of the partitions, i.e., given an application characterized by its class set $\mathcal{C}_A$, we aim to find:
\begin{equation}
    \mathcal{P}^*_A = \argmax\limits_{\mathcal{P} \in B_{\mathcal{C}_A}} \sum\limits_{i=1}^p \alpha_i m_i(\mathcal{P}; f) \label{eq:goal}
\end{equation}
where $B_{C_A}$ denotes the set of all partitions of $C_A$ and $p$ is the number of partitions.

The following is structured as follows. We first discuss prior approaches, and the issues with them. These prior approaches come from a recent review of automated microservice partitioning (\cite{yedida2021lessons}), which studied the effect of hyper-parameter optimization on these approaches. Then, we will present the business case for a new algorithm, and finally, we will discuss our approach.


\begin{table}
\caption{Deep learning: a tutorial.
In this work,
deep learning is used by both
CO-GCN and {\IT}.}\label{dltut} { \small
\begin{tabular}{|p{.98\linewidth}|}\hline\rowcolor{blue!5}
 
   A deep learner is a directed acyclic graph.
   Nodes are arranged into ``layers", which are proper subsets of the nodes. Each node computes a weighted sum of its inputs, and then applies an ``activation function", yielding its output.
 \\
   Deep learning  (DL) (\cite{goodfellow2016deep}) is an extension of prior work on multi-layer perceptrons, where the adjective ``deep'' refers to the use of multiple layers in the network.
  \\\rowcolor{blue!5}
    
    The weights form the parameters of the model, which are learned via \textit{backpropagation} (\cite{rumelhart1986learning}) using the rule $
        \theta = \theta - \eta \nabla_{\theta} \mathcal{L}$
    where $\theta$ represents the parameters of the model, $\eta$ is the \textit{learning rate}, and $\mathcal{L}$ is the \textit{loss function} (described below).
    \\
    A deep learner with $L$ layers produces a prediction $\hat{y}$, and the network learns parameters by using gradient descent to minimize the \textit{loss function} $\mathcal{L}(y, \hat{y})$. This loss function can be arbitrarily designed, as done by the authors of CO-GCN (\cite{desai2021graph}) to suit the specific needs of the application. 
    \\\rowcolor{blue!5}
 For static learning rates, there has been an increased interest in learning rate schedules, where $\eta$ is replaced by a function $\eta(t)$, where $t$ is the iteration. The proposed schedules have both theoretical (\cite{seong2018towards, yedida2021lipschitzlr}) and empirical (\cite{smith2017super, smith2017cyclical}) backing. Generally, however, all the papers agree on the use of non-monotonic learning rate schedules. Specifically, \citet{smith2017cyclical} argues for cyclical learning rates, where the learning rate oscillates between $(\eta_l, \eta_u)$, a preset range for a specified number of cycles. More recently, \citet{smith2019super} proposed a ``1cycle" learning rate policy, where a single cycle is used, and showed this to be effective, but simple. 
\\
    \textit{Dropout} (\cite{srivastava2014dropout}) is a technique which involves removing some nodes with a probability $p$ during training, and adjusting the weights of the model during testing. \citet{srivastava2014dropout} argues that this enforces sparsity in the model weights, which improves the model performance and makes it more robust to noise.
\\\hline
\end{tabular} }
\end{table}

\subsection{FoSCI}

\textbf{FoSCI} (\cite{jin2019service}) uses runtime traces as a data source. They prune traces that are subsets of others, and use hierarchical clustering with the Jaccard distance to create ``functional atoms". These are merged using NSGA-II (\cite{deb2002fast}), a multi-objective optimization algorithm to optimize for the various goals. Since FoSCI relies on a somewhat older optimization algorithm (from 2002~(\cite{deb2002fast})), and all the goals are given equal priority, this can lead to suboptimal results if the different goals have, for example, different scales.

\subsection{Bunch}

\textbf{Bunch} (\cite{mitchell2006automatic}) is based on search techniques such as hill-climbing\footnote{Their tool offers other heuristic-based approaches as well.} to find a partition that maximizes two specific metrics. It is well established that such greedy search algorithms can get stuck in local optima easily (\cite{russell2002artificial}).

\subsection{Mono2Micro}

\textbf{Mono2Micro} (\cite{kalia2020mono2micro, kalia2021mono2micro}) collects runtime traces for different business use cases. Then, they use hierarchical clustering with the Jaccard distance to partition the monolith into a set of microservices. However, as noted by \citet{yedida2021lessons}, this approach takes as input the number of partitions, which different architects may disagree on, or may not know the value of.

\subsection{CO-GCN}

\textbf{CO-GCN} uses the deep learning
technology discussed in Table~\ref{dltut}.
More specifically,
\textbf{CO-GCN} (\cite{desai2021graph}) uses the call graph (built from the code) as input to a graph convolutional neural network (\cite{kipf2016semi}). They develop a custom loss function (defined in~Table~\ref{dltut}) that relates to the metrics being optimized for, and thus, the neural network can be seen as a black-box system that optimizes for Equation \eqref{eq:goal}. However, while their use of a custom loss function tailored to the goals is novel, their approach has several hyper-parameters that a non-expert may not know how to set, and their study did not explore \HPO.

CO-GCN uses an exponential learning rate scheduling policy (see 
Table~\ref{dltut} for details), where the learning decays exponentially. In our approach, we instead use the 1cycle (\cite{smith2019super}) policy, which has more experimental backing.

CO-GCN also uses \textit{dropout} (\cite{srivastava2014dropout}), which involves removing some nodes with a probability $p$ during training, and adjusting the weights of the model during testing. \citet{srivastava2014dropout} argues that this enforces sparsity in the model weights, which improves the model performance and makes it more robust to noise.

CO-GCN is controlled by the settings of
Table \ref{tab:parameters}. Later in this paper, we show that there are several benefits in using hyper-parameter optimization to automatically select good subsets of these hyper-parameters. Note that (a)~all our partitioning methods
(CO-GCN and all the others shown above)   utilize hyper-parameter optimization;
(b)~even better results can be obtained via augmenting HPO with a novel
loss function (and that combination of HPO+loss function is what we call {\IT}).

\subsection{The case for a new algorithm}
\label{sec:why}

The discussion so far shows that there are several approaches that perform automatic microservice partitioning, with different goals, i.e., businesses may be able to choose based on the goals they need, which application to use. However, our industry partners have stressed that these are not widely adopted for several reasons.

Key among these reasons is that these approaches all have \textit{hyper-parameters}. Industry practitioners, having worked with their monolithic systems, may not know how many partitions would be ``ideal" for a microservice architecture. Furthermore, there may be disagreements between different practitioners on the number of microservices (clusters), for example. Moreover, hiring an expert such as a system architect can be expensive, and even then provides no guarantees that just because the architect recommends, say, 4 microservices, that the system will build the microservices envisioned by the architect.

On top of the above, refactoring a monolithic application into microservices is \textit{expensive}. It is a time-consuming process that can take months to complete, with rigorous testing required to ensure that the outward functionality of the system has not been changed. Therefore, it is crucial to large enterprises that if and when such a change is made, that it is done well. What ``well" means can vary from business to business, or even application to application. For example, while a commercial product that faces millions of users might have the non-functional requirement that it be fast, an internal tool may not have such a requirement.

Finally, we bring up the results of the study by \citet{yedida2021lessons}. Specifically, they studied microservice partitioning algorithms, and showed that ``there is no best algorithm". That is, each algorithm that they studied (when tuned with hyper-parameter optimization) was an expert at one of the metrics, but did poorly on the rest. Of course, this does not inspire much confidence in businesses who might be open to adopting these automated systems, and whose requirements may be constantly changing.

\subsection{{\IT}}
\label{sec:method}

Summarizing the approaches in prior work, we note that they suffer from the following limitations:
\begin{enumerate}[(a)]
    \item They have hyper-parameters that practitioners may not understand or know how to set.
    \item They rely on techniques that can easily get stuck in suboptimal, local minima.
    \item They treat all the metrics as having equal priority, when it may not be the case (e.g., some feature scaling may be needed, or the business priorities are different).
\end{enumerate}
To address the limitations, we propose an extension
to the  CO-GCN (\cite{desai2021graph}) deep learner. 
{\IT} uses  an  \HPO technique that is known to deal with local optima better, and (b) a novel reweighting of the metrics based on data.

When algorithms produce a highly variable output, then
\HPO, described in \S \ref{sec:hpo},
can be used to automatically learn the control settings
 that minimize the variance.
 Specifically, we use the fact that a \HPO algorithm, in searching for optimal parameters, must test different hyper-parameter configurations, each of which produces a different result. The results are aggregated together to form a  frontier, from which we can pick the ``best" sample. To do so, we notice that simply picking a model based on one metric sacrifices the performance in others. Moreover, using a sum of all metrics has the disadvantages of (a) different scales for different metrics (b) learners doing well in the ``easy" metrics but not in the others (c) ignoring correlations among the metrics. Therefore, we design a custom \textit{loss function} to choose the best sample. 

For further details on all the concepts in the last
two paragraphs, see the rest of this section.

\subsubsection{Feature Engineering}
\label{sec:method:data}

{\IT}  uses the deep learning
technology discussed in Table~\ref{dltut}.


We format our datasets into the form required by a graph convolutional network (\cite{kipf2016semi}) as suggested in CO-GCN (\cite{desai2021graph}). A graph convolutional network models a graph as a pair of matrices $(A, X)$, where $A$ is the (binary) adjacency matrix and $X$ is the feature matrix. Our graph is characterized by nodes corresponding to classes, and edges corresponding to function calls between two classes. If a class is not used by any of the APIs published by the software, then it is removed from the graph. The adjacency matrix is trivially defined as $M_{ij} = 1$ if an edge exists between the vertices $i, j$, and 0 otherwise. 

For the feature matrix, we follow the approach of suggested in CO-GCN (\cite{desai2021graph}). Specifically, we first define the \textit{entry point matrix} $E \in \{0, 1\}^{|V| \times |P|}$  (where $V$ is the set of classes, and $P$ is the set of \textit{entry points}). Entry points refer to APIs published by a software, each potentially for different functions. Then, for each such entry point (i.e., API), we consider the set of classes invoked in its execution. We consider the entry $E_{ij} = 1$ if class $i$ is part of the execution trace of entry point $j$, and 0 otherwise. Additionally, we consider the \textit{co-occurrence matrix} $C \in \{0, 1\}^{|V| \times |V|}$ such that $C_{ij} = 1$ if both classes $i, j$ occur in the same execution trace, and 0 otherwise. Finally, we define the \textit{dependence matrix} $D \in \{0, 1\}^{|V| \times |V|}$ as $D_{ij} = D_{ji} = 1$ if class $i$ inherits from class $j$ or vice versa, and 0 otherwise. The feature matrix $X \in \mathbb{R}^{|V| \times (|P| + 2|V|)}$ is the concatenation of $E, C, D$ (in that order), and is then row-normalized.

CO-GCN also uses \textit{dropout} (\cite{srivastava2014dropout}), which involves removing some nodes with a probability $p$ during training, and adjusting the weights of the model during testing. \citet{srivastava2014dropout} argues that this enforces sparsity in the model weights, which improves the model performance and makes it more robust to noise.

\subsubsection{Hyper-parameter Optimization}
\label{sec:hpo}
Hyper-parameter optimization is the systematic process of finding an optimal set of hyper-parameters for a model. In the specific case of optimizing CO-GCN, those parameters are shown in  Table~\ref{tab:parameters}.

Several approaches for \HPO now exist. Of note is the work by \citet{bergstra2011algorithms}\footnote{As of August 2021, this paper has 2,800 citations.} which discusses three different approaches. 
In this paper, we use a newer, widely used \HPO algorithm called Tree of Parzen Estimators (TPE) (\cite{bergstra2011algorithms}). TPE divides data points seen so far into \textit{best} and \textit{rest}, each of which are modeled by a Gaussian distribution. New candidates are chosen so that they are highly likely to be in the \textit{best} distribution. Evaluating these candidates adds to the data points, and the algorithm continues. This algorithm was open-sourced by its authors in a package called \textit{hyperopt} (\cite{bergstra2013hyperopt}). Therefore, in this paper, whenever we say ``hyperopt", we mean TPE as implemented by this package.

\subsubsection{Loss Function}
\label{sec:method:overall}

Table~\ref{dltut} discussed {\em loss functions} $\mathcal{L}$ that offers
weights to the feedback seen by  the learner during the inner-loop of the learner process.
Numerous researchers report that augmented loss functions can enhance reasoning:
\bi
\item \citet{ryu2016effective} used reweighted Naive Bayes for cross-project defect prediction;
\item In a similar approach, a reweighted Naive Bayes was used by \citet{arar2017feature} for defect prediction and \citet{yedida2021value} explored weighted loss functions in deep learners for defect prediction. 
\item
In the AI literature, \citet{lin2017focal} propose a ``focal loss" function that uses an exponential weighting scheme. 
\ei
To the best of our knowledge, augmenting the loss function has not been previously
attempted for improving microservice partitions. 

\begin{figure*}
    \centering
    \includegraphics[width=\textwidth]{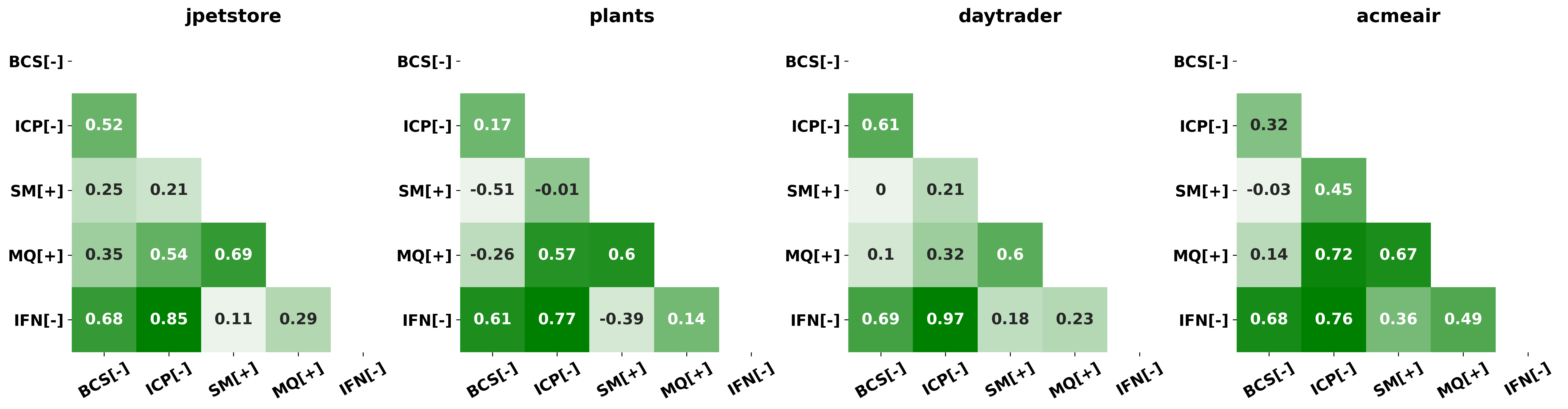}
    \caption{Spearman $\rho$ among metrics from 1,000 runs.}
    \label{fig:corr}
\end{figure*}

One challenge with designing a  loss function is how to   obtain 
the required weights. In {\IT}, we 
first run the algorithm 1,000 times to generate a set of metrics shown in Table~\ref{tab:metrics}. Then, we check the correlation among the metrics using the Spearman correlation coefficients. If any two metrics have strong correlations we prefer to keep one to remove the redundant metrics for the optimization.
Figure \ref{fig:corr} indicates the correlations among the metrics. Clearly, some of the metrics are highly correlated. Therefore, we use a reduced set of metrics for evaluating our approaches. We set a threshold of 0.6 to prune the set of metrics. We observe that MQ has higher correlations with other metrics than SM. Since IFN and ICP are highly correlated across all datasets, we arbitrarily choose ICP. This leads to the final set of metrics: BCS, ICP, and SM. Note that to ensure that we do not generate ``dust" (individual classes in a partition) or ``boulders" (monolithic partitions), we also include NED in the final metric set (for which we assigned a static weight of 0.2).  

Based on these three metrics, we choose a loss function to pick one model from the   frontier of best solutions. This loss function is

\begin{equation}
    \mathcal{L}_H(P) = \sum_i w_i m_i(P) \label{eq:loss}
\end{equation}

where $w_i$ are \textit{weights} assigned to each of the metrics $m_i$, and $P$ is the partition. 
Additionally, we divide the BCS weight by 10 (for an overall weight of 0.1) to normalize its value to the same order of magnitude as the other metrics. 
Finally, we add in a term for the NED metric, so that we do not generate ``dust" or ``boulders" (e.g., a monolith with 20 classes is not partitioned into 19 and 1). 

Next, we change the clustering method used in CO-GCN (\cite{desai2021graph}). Rather than using the k-means clustering, which assumes globular clusters, we use spectral clustering, which can produce more flexible cluster shapes~(\cite{yedida2021beginning}). We also change the cluster loss function to
\begin{equation}
    \mathcal{L}_{clus} = \sum\limits_i \min\limits_j \left| x_i - x_j \right|
\end{equation}
where $x_i$ and $x_j$ are data points.

Next, we update the learning rate scheduling policy to the ``1cycle" policy (\cite{smith2017cyclical,smith2019super}), which has been shown to lead to faster convergence (see Table~\ref{dltut}  for details). We update the learning rate every 200 steps.

Finally, we change all the activation functions in the network to ReLU ($f(x) = \max(0, x)$) (\cite{nair2010rectified}), except the last layer (which we leave unchanged to $f(x) = x$). Specifically, like \citet{desai2021graph}, we use an encoder-decoder architecture with 2 layers per each layer; each layer being a graph convolution as defined by \citet{kipf2016semi}. Therefore, for an input \textit{feature matrix} $X$, we consider
\begin{align}
    E \gets& \mathrm{ReLU}(\hat{A}(\mathrm{ ReLU}(\hat{A}XW_1))W_2) \label{eq:encoder} \\
    D \gets& \mathrm{ReLU}(\hat{A}(\mathrm{ ReLU}(\hat{A}EW_3))W_4) \label{eq:decoder}
\end{align}
Algorithm \ref{alg:cogcnv2} shows the overall approach. In the algorithm 
{\color{red}the lines in red\color{black}} 
(19-22,25) are from
     the CO-GCN algorithm (and all else is our extension).
     We first collect correlation statistics using a Monte Carlo-style data collection (lines 1-4). Then we set weights (lines 4-7) and run hyperopt for 100 iterations (lines 10-15), while collecting the results of each run. The weighted loss function then picks the ``best" candidate in line 16, and returns the optimal hyper-parameters in line 17. The actual model training itself is delegated to \texttt{CO-GCN-modified}. 

 We form the graph Laplacian in lines 18-20, and pretrain (for 1 epoch) the encoder (see Eq. \eqref{eq:encoder}) and decoder (see Eq. \eqref{eq:decoder}). We initialize the clusters in line 23, but use spectral clustering instead. Then, we run the actual training for 300 epochs.

In summary, our new method {\IT} differs from CO-GCN in the following ways:

\begin{algorithm}[!htbp]
\small
    \SetAlgoLined
    \DontPrintSemicolon
    \SetKwInOut{KwInput}{Input}
    \SetKwInOut{KwOutput}{Output}
    \SetKwProg{myalg}{Algorithm}{}{}
    
    \myalg{{\IT}}{
        \KwInput{graph dataset $D$}
        \KwOutput{optimal hyper-parameters $\theta^*$}
        
        \For{1,000 iterations}{
            run \texttt{CO-GCN-modified($D$)} with random settings and collect metrics\;
        }
        $w_{SM} \gets -1 / corr(SM, MQ)$ \tcp*{initialize weights}
        $w_{ICP} \gets 1 / corr(BCS, ICP)$\;
        $w_{NED} \gets 0.2, w_{BCS} \gets 0.1$\;
        $\theta = \phi$ \tcp*{initialize our history to empty set}
        $M = \phi$ \tcp*{initialize results to empty}
        \For{i=1..100}{
            $\theta_i \gets$ hyperopt($\theta$)\;
            $M_i \gets$ \texttt{CO-GCN-modified}($D; \theta_i$)  \tcp*{run the model}
            append $\theta_i$ to $\theta$ \tcp*{log hyper-params}
            append $M_i$ to $M$ \tcp*{log results}
        }
        
        $\theta^* = \argmin\limits_\theta \sum\limits_{j=1}^{4} w_j M_{j}$\; 
        \Return $\theta^*$\;
    }
    
    \myalg{CO-GCN-modified}{
        \KwInput{graph dataset $D = (A, X)$}
        \KwOutput{partitioned microservices}
        \color{red}{ 
        \tcp{compute the graph Laplacian}
        $\tilde{A} \gets A + I$\;
        define diagonal matrix $\tilde{D}$ such that $\tilde{D}_{ii} = \sum\limits_{j} \tilde{A}_{ij}$\;
        $\hat{A} \gets \tilde{D}^{-\frac{1}{2}}\tilde{A}\tilde{D}^{-\frac{1}{2}}$\;
        
        pretrain the encoder and decoder using \eqref{eq:encoder} and \eqref{eq:decoder} respectively\; }
        \color{black}{
         initialize the clusters using spectral clustering on embeddings}\;
        train the network $(E, D)$ for 300 epochs using Alg. 1 lines 5-9 from (\cite{desai2021graph}) and 1cycle (\cite{smith2019super})\;
        \color{red}{get embeddings using \eqref{eq:encoder}}\;
        \color{black}{cluster the embeddings using spectral clustering}\;
        \Return{clusters}
    }
    
    \caption{{\IT}. Note {\em CO-GCN-modified}
     is a sub-routine containing our changes to
     the original CO-GCN system. 
     The {\color{red}the lines in red\color{black}} (18-21,24) are from
     the original CO-GCN algorithm (and all else is our extension).
    }
    \label{alg:cogcnv2}
\end{algorithm}

\begin{itemize}
    \item We use a better learning rate scheduling policy from deep learning literature (discussed in Table~\ref{dltut}).
    \item We update the activation functions used in the neural network (also see   Table~\ref{dltut}).
    \item We run \HPO and choose from the Pareto frontier using a custom loss function  over a reduced space of metrics (discussed in \S \ref{sec:hpo}).
    \item We update the clustering method employed, and the corresponding term in the deep learner loss function (see   \S \ref{sec:method:overall}).
\end{itemize}

\section{Experimental Design}
\label{sec:setup}

In this section, we detail our experimental setup for the various parts of our research. Broadly, we follow the same experimental setup as \citet{yedida2021lessons}. However, that paper concluded that different optimizers performed differently across datasets and metrics, and that there was no one winning algorithm. In this paper, using the method described in \S \ref{sec:method}, we show that our approach wins most of the time across all our datasets and metrics.

\subsection{Case studies}
\label{sec:cases}

We use four open source projects to evaluate our approach against prior work. These are \textit{acmeair}\footnote{\url{https://github.com/acmeair/acmeair}} (an airline booking application), \textit{daytrader}\footnote{\url{https://github.com/WASdev/sample.daytrader7}} (an online stock trading application), \textit{jpetstore}\footnote{\url{https://github.com/mybatis/jpetstore-6}} (a pet store website), and \textit{plants}\footnote{\url{https://github.com/WASdev/sample.plantsbywebsphere}} (an online web store for plants and pets). These applications are built using Java Enterprise Edition (J2EE), and common frameworks such as Spring and Spring Boot.

In Table \ref{tab:datasets}, we show statistics about the datasets (number of classes and methods) and the runtime traces that we used (number of traces, and their class and method coverage). 
While the coverage here may seem low, we note that applications tend to have a significant amount of ``dead'', or unreachable code. For example, \citet{brown1998antipatterns} found between 30 to 50\% of an industrial software system was dead code. \citet{eder2012much} found that for an industrial software system written in .NET, 25\% of method genealogies were dead. Therefore, we were not worried about the coverage seeming low.

\begin{table}
    \centering
    \caption{Statistics about the datasets used in this study.}
    \label{tab:datasets} { \footnotesize
    \begin{tabularx}{.8\linewidth}{lllLLL}
        \toprule
        \textbf{Application} & \textbf{\#classes} & \textbf{\#methods} & \textbf{\#Runtime traces} & \textbf{Class coverage} & \textbf{Method coverage} \\
        \midrule
        acmeair & 33 & 163 & 11 & 28 (84\%) & 108 (66\%) \\
        daytrader & 109 & 969 & 83 & 73 (66\%) & 428 (44\%) \\
        jpetstore & 66 & 350 & 44 & 36 (54\%) & 236 (67\%) \\
        plants & 37 & 463 & 43 & 25 (67\%) & 264 (57\%) \\
        \bottomrule
    \end{tabularx} }
\end{table}

\subsection{Hyper-parameter Optimization}
\label{sec:setup:hpo}

We use the Tree of Parzen Estimators (TPE) (\cite{bergstra2011algorithms}) algorithm from the hyperopt (\cite{komer2014hyperopt}) package for \HPO. As discussed in \S \ref{sec:method}, we use $\mathcal{L}_H(P)$ from \eqref{eq:loss} to guide hyperopt towards an optimal set of hyper-parameters. Table \ref{tab:parameters} lists the hyper-parameters that we tune, along with their ranges. We run the hyper-parameter optimizer for 100 iterations. We train for 300 epochs using an initial learning rate of 0.01.

\subsection{Baselines}
\label{sec:setup:baselines}

Following our literature review, we use the baselines shown in Table \ref{tab:parameters}. All these approaches, except Mono2Micro and CO-GCN, were either open-source, or re-implemented by us. For a fair evaluation, we tune each algorithm with hyperopt for 100 iterations (which is the same as our approach). We use the same loss function $\mathcal{L}_H(P)$ shown in \eqref{eq:loss}, but set all the weights to either 1 (for metrics we wish to minimize), or -1 (for metrics we wish to maximize), since the correlation-based weights are a novel idea for our approach.

\subsection{Metrics}
\label{sec:setup:metrics}

For a fair comparison with prior work, we must compare using the same set of metrics. We choose a total of five metrics to evaluate our core hypothesis that hyper-parameter tuning improves microservice extraction algorithms. These are detailed below. These metrics have been used in prior studies, although different papers used different set of metrics in their evaluations. For fairness, we use metrics from all prior papers. These metrics evaluate different aspects of the utility of an algorithm that might be more useful to different sets of users (detailed in the RQ1), e.g., BCS evaluates how well different business use cases are separated across the microservices.

{\em Inter-partition call percentage (ICP)} (\cite{kalia2020mono2micro}) is the percentage of runtime calls across different partitions. For lower coupling, \textit{lower} ICP is better.

{\em Business context sensitivity (BCS)} (\cite{kalia2020mono2micro}) measures the mean entropy of business use cases per partition. Specifically, 

\[
    BCS = \frac{1}{K} \sum\limits_{i=1}^K \frac{BC_i}{\sum_j BC_j} \log_2 \left( \frac{BC_i}{\sum_j BC_j} \right)
\]
where $K$ is the number of partitions and $BC_i$ is the number of business use cases in partition $i$. Because BCS is fundamentally based on entropy, \textit{lower} values are better.

{\em Structural modularity (SM)}, as defined by \citet{jin2019service}, combines cohesion and coupling, is given by

\begin{align*}
    SM &= \frac{1}{K} \sum_{i=1}^n \frac{coh_i}{N_i^2} - \frac{1}{N_i(N_i-1)/2} \sum_{i < j < K} coup_{i,j} \\
    coh_i &= \frac{2}{m(m-1)} \sum\limits_{\substack{c_i, c_j \in C \\ i < j}} \sigma(c_i, c_j) \\
    coup_{i, j} &= \frac{\sum\limits_{\substack{c_1 \in C_i \\ c_2 \in C_j}} \sigma(c_1, c_2)}{\lvert C_i \rvert \lvert C_j \rvert}
\end{align*}

where $N_i$ is the number of classes in partition\footnote{Note that we use the term ``partition" to refer to both the set of all class subsets, as well as an individual subset, but it is typically unambiguous.} $i$, $K$ is the number of partitions, $coh_i$ is the cohesion of partition $i$, and $coup_{i,j}$ is the coupling between partitions $i$ and $j$, $c_i$ refers to a class in the subset, $C_i, C_j$ are two partitions, $\sigma$ is a general similarity function bounded in $[0, 1]$, and the $i < j$ condition imposes a general ordering in the classes. Higher values of SM are better. 


{\em Modular quality (MQ)} (\cite{mitchell2006automatic}), coined by \citet{mitchell2006automatic}, is defined on a graph $G = (V, E)$ as

\begin{align*}
    MQ &= \frac{2\mu_i}{2\mu_i+\epsilon} \\
    \mu_i &= \sum\limits_{(e_1, e_2) \in E} \mathbbm{1}(e_1 \in C_i \land e_2 \in C_i) \\
    \epsilon &= \sum\limits_{\substack{(e_1, e_2) \in E \\ i \neq j}} \mathbbm{1}(e_1 \in C_i \land e_2 \in C_j)
\end{align*}

where $\mathbbm{1}$ is the indicator function, and $C_i, C_j$ are clusters in the partition. Higher values of MQ are better.

The {\em interface number (IFN)} (\cite{mitchell2006automatic}) of a partition is the number of interfaces needed in the final microservice architecture. Here, an interface is said to be required if, for an edge between two classes in the runtime call graph, the two classes are in different clusters.

Finally, the {\em non-extreme distribution (NED)} is the percentage of partitions whose distributions are not ``extreme" (in our case, these bounds were set to min=5, max=20).

However, as shown in Figure \ref{fig:corr}, some of these metrics are highly correlated with others. Therefore, to avoid bias in the evaluation and the loss function (i.e., if metrics $M_1$ and $M_2$ are correlated, and optimizer $O$ performs best on $M_1$, it likely performs best on $M_2$ as well), we use a subset of these metrics. Specifically, we prune the metric set as discussed in \S \ref{sec:method:overall}.
Note that across all our figures and tables, a ``[-]" following a metric means lower values are better, and a ``[+]" following a metric means that higher values are better.

\subsection{Statistics}
\label{sec:setup:stats}

For comparing different approaches, we use statistical tests due to the stochastic nature of the algorithms. According to standard practice (\cite{ghotra2015revisiting}), we run our algorithms 30 times to get a distribution of results, and run a Scott-Knott test as used in recent work (\cite{agrawal2019dodge,yedida2021value}) on them. The Scott-Knott test is a recursive bi-clustering algorithm that terminates when the difference between the two split groups is insignificant. The significance is tested by the Cliff's delta using an effect size of 0.147 as recommended by \citet{hess2004robust}. Scott-Knott searches for split points that maximize the expected value of the difference between the means of the two resulting groups. Specifically, if a group $l$ is split into groups $m$ and $n$, Scott-Knott searches for the split point that maximizes

\[
    \mathbb{E}[\Delta] = \frac{|m|}{|l|}\left( \mathbb{E}[m] - \mathbb{E}[l] \right)^2 + \frac{|n|}{|l|}\left( \mathbb{E}[n] - \mathbb{E}[l] \right)^2
\]

where $|m|$ represents the size of the group $m$.

The result of the Scott-Knott test is \textit{ranks} assigned to each result set; higher the rank, better the result. Scott-Knott ranks two results the same if the difference between the distributions is insignificant.

\section{Results}
\label{sec:results}
   

\noindent
\underline{\textbf{RQ1:}}  {\em How prevalent is hyper-parameter brittleness in automated microservice partitioning?}

Here, we ran  the approaches from Table \ref{tab:parameters}, 30 times each,  then compared results from those treatments with the statistical methods of 
\S\ref{sec:setup:stats}.
The median results are shown in Table \ref{tab:results:scores}
and the statistical analysis is shown  in Table \ref{tab:results:sk}.

In those results, we see that across different datasets and metrics, those ranks vary widely, with little consistency across the space of datasets and metrics.
For example:
\bi
\item For jpetstore, we see that the best algorithms for the three metrics are Mono2Micro, CO-GCN, and MEM respectively; 
\item
For acmeair, the best algorithms are Mono2Micro, and Bunch; 
\item
For plants, the best algorithms are FoSCI, CO-GCN, and MEM;
\item
For daytrader, the best are Mono2Micro, CO-GCN, and Bunch.
\ei
Hence we say: 
\begin{formal}
    \noindent
    For all the data sets and goals  studied here, there is widespread performance brittleness.
\end{formal}
As to why this issue has not been reported before in the literature,
we note that much of the prior work was focused on one algorithm exploring one case study. To the best of our knowledge, this is first to perform a comparison across the same datasets and metrics and also propose a novel approach.  

\begin{table}[!t]
    \centering
    \caption{Results on all datasets. The top row shows our median performance scores over 30 runs, while the bottom row shows the Scott-Knott ranks (lower is better). \fcolorbox{black}{fill}{Gray} cells indicate the best result according to the Scott-Knott test. All algorithms were tuned by hyperopt.}
    \label{tab:results} { \scriptsize
    \begin{subtable}{\textwidth}
        \centering
        \caption{Median performance scores over 30 runs}
        \label{tab:results:scores} 
        \begin{tabular}{l|lll|lll|lll|lll}
            \toprule
            \multirow{2}{*}{\textbf{Algorithm}} & \multicolumn{3}{c}{jpetstore}  & \multicolumn{3}{c}{acmeair} & \multicolumn{3}{c}{plants} & \multicolumn{3}{c}{daytrader} \\
            \cmidrule{2-13}
            & \textbf{BCS [-]} & \textbf{ICP [-]} & \textbf{SM [+]} & \textbf{BCS [-]} & \textbf{ICP [-]} & \textbf{SM [+]} & \textbf{BCS [-]} & \textbf{ICP [-]} & \textbf{SM [+]} & \textbf{BCS [-]} & \textbf{ICP [-]} & \textbf{SM [+]} \\
            \midrule
            Bunch & 2.43 & 0.48 & 0.21 & 1.67 & 0.55 & \gray{0.17} & 2.29 & 0.56 & 0.22 & 1.86 & 0.57 & \gray{0.27} \\
            Mono2Micro & \gray{1.67} & 0.53 & 0.05 & \gray{1.58} & 0.4 & 0.06 & 2.49 & 0.39 & 0.07 & \gray{1.58} & 0.4 & 0.07 \\
            CO-GCN & 2.79 & 0.27 & 0.2 & 1.89 & 0.47 & 0.15 & 2.46 & 0.35 & 0.24 & 2.09 & 0.05 & 0.13 \\
            FoSCI & 1.99 & 0.66 & 0.05 & 2.08 & 0.54 & 0.07 & 1.94 & 0.45 & 0.07 & 2.04 & 0.48 & 0.07 \\
            MEM & 2.74 & 0.48 & 0.22 & 2.13 & 0.69 & 0.03 & 2.04 & 0.42 & 0.25 & 2 & 0.59 & 0.09 \\
            \midrule
            \textit{{\IT}} & 2.35 & \gray{0.06} & 0.2 & \gray{1.41} & \gray{0.18} & 0.16 & \gray{1.87} & \gray{0.02} & \gray{0.27} & 1.74 & \gray{0.03} & 0.17 \\
            \bottomrule \\
        \end{tabular}
    \end{subtable}
    \begin{subtable}{\textwidth}
        \centering
        \caption{Scott-Knott ranks. For all metrics, \textit{lower} is \textit{better}.}
        \label{tab:results:sk}
        \begin{tabular}{l|lll|lll|lll|lll}
            \toprule
            \multirow{2}{*}{\textbf{Algorithm}} & \multicolumn{3}{c}{jpetstore}  & \multicolumn{3}{c}{acmeair} & \multicolumn{3}{c}{plants} & \multicolumn{3}{c}{daytrader} \\
            \cmidrule{2-13}
            & \textbf{BCS [-]} & \textbf{ICP [-]} & \textbf{SM [+]} & \textbf{BCS [-]} & \textbf{ICP [-]} & \textbf{SM [+]} & \textbf{BCS [-]} & \textbf{ICP [-]} & \textbf{SM [+]} & \textbf{BCS [-]} & \textbf{ICP [-]} & \textbf{SM [+]} \\
            \midrule
            Bunch & 4 & 3 & 2 & 2 & 4 & \gray{1} & 4 & 6 & 4 & 3 & 5 & \gray{1} \\
            Mono2Micro & \gray{1} & 5 & 5 & \gray{1} & 2 & 4 & 6 & 3 & 5 & \gray{1} & 3 & 6 \\
            CO-GCN & 6 & 2 & 2 & 3 & 3 & 3 & 5 & 2 & 3 & 6 & 2 & 3 \\
            FoSCI & 2 & 6 & 4 & 4 & 4 & 5 & 2 & 5 & 6 & 5 & 4 & 5 \\
            MEM & 5 & 4 & 1 & 5 & 5 & 6 & 3 & 4 & 2 & 4 & 6 & 4 \\
            \midrule
            \textit{{\IT}} & 3 & \gray{1} & 3 & \gray{1} & \gray{1} & 2 & \gray{1} & \gray{1} & \gray{1} & 2 & \gray{1} & 2 \\
            \bottomrule
        \end{tabular}
    \end{subtable} }
\end{table}


\noindent
\underline{\textbf{RQ2:}} {\em  Is hyper-parameter optimization enough to curb the aforementioned optimizer brittleness?}

Here we check of brittleness can be solved by tuning the partitioning methods.

Since all the algorithms in our study were tuned by hyperopt, the results of Table \ref{tab:results} show that even post-tuning, there is a large brittleness problem.   Accordingly, we say:

\begin{formal}
    \noindent
    Hyper-parameter optimization does not suffice to fix performance brittleness.
\end{formal}

 Since hyper-parameter optimization is not enough to tame
 brittleness,  we 
  ask:

\noindent
\underline{\textbf{RQ3:}} {\em How else might we  fix the aforementioned brittleness problem?}

When we added the novel loss function (discussed above in 
\S\ref{sec:method:overall}), we obtained the results shown in the 
last line of the two tables of Table \ref{tab:results:sk}. When measured in terms of
ranks seen amongst the different populations, the last line of Table \ref{tab:results:sk} is most insightful. Here we see that in general, {\IT} performs better compared to all other optimizers across all our case studies.   Hence we say:



\begin{formal}
    \noindent
    Weighted losses together with {\HPO} fix the brittleness problem.
\end{formal}

\begin{figure}
    \centering
        \includegraphics[width=.6\linewidth]{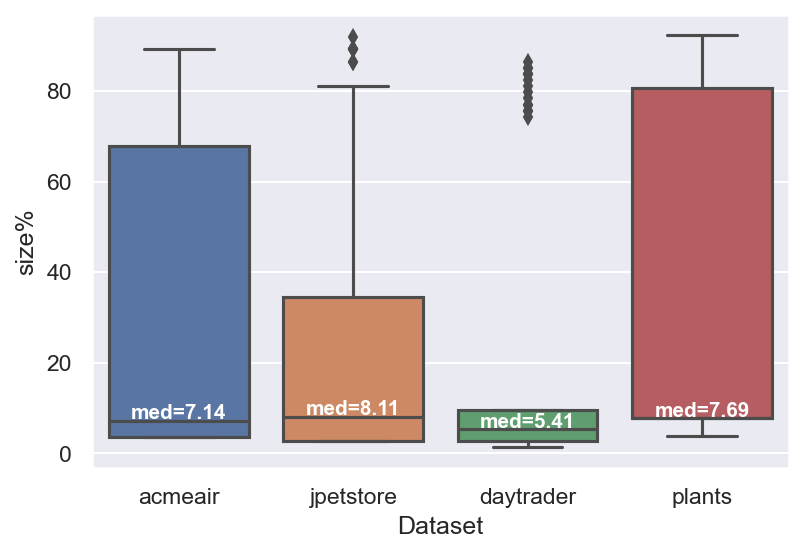}
        \label{fig:boxplot:sizes}
    \caption{Analysis of partitions generated by our approach. Distribution of class size percentages across datasets (median results).}
 \label{fig:boxplot:lens}
 \vspace{-6pt}
\end{figure}

\noindent
\underline{\textbf{RQ4:}} {\em Do we generate ``dust" or ``boulders"?}

Two anti-patterns of partitioning are ``dust'' (generating too many partitions) and ``boulders''  (generating too few partitions).

To analyze this, Figure~\ref{fig:boxplot:lens} shows the distribution
of partition sizes seen in 30 random samples.   From that figure we say that our methods usually generate small number of partitions. Also, since we are usually dealing with classes ranging from 33 in acmeair to 109 in daytrader, 
Figure~\ref{fig:boxplot:lens} tells us that we are
not generating partitions with only one item per partition (for example, on plants, the median size is $5.4\% \times 111 = 6$ classes).
Hence we say: 

\begin{formal}
    Our partitions are neither ``boulders'' nor ``dust''.
\end{formal}

\section{Discussion}
\label{sec:broader}

\subsection{Lessons Learned}

\citet{yedida2021lessons} listed four lessons learned from their study.
Our results suggest that some of those lessons now need to be revised.

\begin{enumerate}[leftmargin=*]
    \item We fully concur with Yedida et al. when they said ``{\em do not use   partitioning
    methods   off-the-shelf since these
    these tend to have subpar results}. That said, currently we are planning to package (in Python) the {\IT} system. Once that is available then
    Algorithm~1 should accomplish much of the adjustments required
    to commission {\IT} for a new domain. 
    
    \item Yedida et al. said ``{\em the choice of tuning algorithm is not as important as the decision to tune. The random and hyperopt optimizers performed differently on different metrics and datasets, but in a broader view, shared the same number of highest statistical 
    rankings. Therefore, the decision to tune is important, but the choice of algorithm may not be as important.}''. We disagree since, as shown by 
    Table \ref{tab:results:sk}, our methods demonstrate overall
    better results (as indicated by the statistical ranks).
      \item 
      Yedida et al. said  ``{\em More research is required on tuning.   While  our  results  shows  that  tuning  is  better than  non-tuning,  we  also  show  that  no  single  approach is  a  clear  winner  across  multiple  datasets  or  multiple performance goals.}''.
      We concur but add that {\IT} seems to resolve much of their pessimism about the state of the art in hyper-parameter optimization.
    \item      Yedida et al. said ``{\em There is no best partitioning algorithm...  there is no one best algorithm across all   metrics.}''
    We agree, but only  partially.
    The last row of Table \ref{tab:results:sk} shows that even our
    preferred method {\IT} does not always obtain top rank. That said,
    {\IT} often performs as well or better than anything else.   
\end{enumerate}

\subsection{On the value of dynamic traces}

It is notable that one of the prior approaches we choose to compare against is Bunch and its hill-climbing algorithm. Several years ago, an extensive empirical study by \citet{garcia2013comparative} demonstrated that Bunch performs poorly in identifying meaningful abstractions in real-world systems it analyzes. However, we still chose Bunch as a comparative system, since that study used static attributes as the input to Bunch, while our inputs come from dynamic, runtime traces, which we believe makes it meaningfully different. That is, we use the results of \citet{garcia2013comparative} as an indication that static traces may not be sufficient for the microservice partitioning problem, and this study explores dynamic traces instead.
 
\subsection{Business Implications}

At a higher level, hyper-parameter tuning is akin to searching for various options and guessing which one is the best; weighted loss functions directly encode business goals in this process, making it a more focused effort. For example, changing business goals can be trivially implemented in our framework, since the weighted loss function used in the hyper-parameter tuning is based on the metrics (i.e., no additional mathematical derivations need to be done). 

Importantly, business users can (a) make a choice among many from the frontier of best solutions of a size that they can choose, with the understanding that higher sizes mean more options but takes slightly longer (b) see the trade-offs between different architectural design choices directly in terms of metrics (c) easily choose new metrics if they see fit. That is, our approach provides businesses with flexibility and transparency into the working of the model. While the deep learner itself is a black box, end users typically do not care about the internal workings of the model; only in interpretable ways to do better in their goals (this was studied by \citet{chen2019predicting}).

For businesses, we offer the following advantages:
\begin{itemize}
    \item \textbf{Stability:} Business can be assured that our approach will perform well across any dataset, no matter the metric used. This stability is important for time-constrained businesses who need some guarantees of good results before using a new approach.
    \item \textbf{Performance:} Our approach achieved state-of-the-art performance across different and uncorrelated metrics on four open-source datasets. This high performance inspires confidence in our approach for businesses looking to adopt an automated system for the architectural change.
    \item \textbf{Openness:} Our code is fully open-source, but also modular. That is, businesses are not limited to our specific approach (which builds upon CO-GCN and uses hyperopt); rather, they are free to use our techniques to build upon any existing infrastructure that they may have (e.g., IBM Mono2Micro).
\end{itemize}

\subsection{Research Implications}
The ideas of this paper extend to other tasks and domains. In this section, we elaborate on the specific ideas and their broader utility.

Our approach is a general method that can be applied to any dataset, since the components themselves can be changed. For example, a different hyper-parameter optimizer such as Optuna (\cite{akiba2019optuna}) can be used instead of hyperopt, and a different weighting mechanism can be chosen instead of our correlation-based weights. Finally, a different set of preprocessing can be used. Therefore, a generalized version of our approach is:

\begin{enumerate}[(1)]
    \item \textbf{Feature extraction.} Different features sets can be extracted from code. For example, one might use code2vec \citet{alon2019code2vec}, which transforms code snippets to fixed-length vectors.
    \item \textbf{Hyper-parameter optimization.} Any hyper-parameter optimization approach can be plugged in here, such as random sampling, TPE, DODGE, etc.
    \item \textbf{Loss-based selection.} Any loss function can be applied to the   frontier of best solutions to select a configuration. Moreover, the loss function of the learner itself can be modified, as done in CO-GCN.
\end{enumerate}

From a research standpoint, we offer the following:
\begin{itemize}
    \item \textbf{Advancing the state-of-the-art:} Our approach consistently outperforms all the prior state-of-the-art approaches we tested against across three different metrics, on four datasets.
    \item \textbf{Modular approach:} Our approach can be adapted by changing the base algorithm (CO-GCN) and hyper-parameter optimization algorithm (hyperopt) and used on any dataset as discussed above.
    \item \textbf{Documenting the success of weighted losses:} Our paper adds to the body of literature that documents the success of using weighted losses for different problems, possibly motivating future work to also use them. More generally, our idea can be applied to any multi-objective optimization problem, using the general method from our paper: produce a   frontier of
    best solutions (through hyper-parameter optimization, swarm optimization, etc.) and use a weighted loss to choose the best candidate. This idea has been used implicitly in the deep learning field. In particular, \citet{santurkar2018does,li2017visualizing} show that a smoother loss function surface is beneficial for optimization. In the case of \citet{santurkar2018does}, the adding of ``skip-connections" in the deep learner modify its loss function, making it smoother and therefore easier to optimize in. This idea has also been used by \citet{chaudhari2019entropy}, who change the loss function to shape it into a more amenable form.
\end{itemize}

\section{Threats to Validity}
\label{sec:threats}

\noindent\textbf{Sampling bias:} With any data mining paper, it is important to discuss sampling bias. Our claim is that by evaluating on four different open-source projects across different metrics, we mitigate this. Nevertheless, it would be important future work to explore this line of research over more data.

\noindent\textbf{Evaluation bias:} While we initially considered comparing across all the metrics we found in the literature, we noticed large correlations between those metrics. Therefore, to reduce the effect of correlations, we chose a subset of the metrics and evaluated all the approaches across those. Moreover, in comparing the approaches, we tuned all of them using the same {\HPO} algorithm, for the same number of iterations.

\noindent\textbf{External validity:} We tune the hyper-parameters of the algorithm, removing external biases from the approach. Our baselines are also tuned using hyperopt for the same number of iterations.

\textbf{Internal validity:} All algorithms compared were tuned using hyperopt. Because our approach involves using weighted losses and other tweaks to CO-GCN, these w   ere not applied to the other algorithms.

\section{Conclusion}
\label{sec:conclusion}

In this paper, we presented a systematic approach for achieving state-of-the-art results in microservice partitioning. Our approach consists of hyper-parameter optimization and the use of weighted losses to choose a configuration from the   frontier of best solutions. 

Broadly, the lesson from this work is:

\begin{formal}
    \noindent
   At least for microservice partitioning,  weighted loss functions can work together with tuning to achieve superior results.
\end{formal}

We first analyzed the existing state-of-the-art. Through this review, we noticed (a) highly correlated, and therefore redundant, metrics used in the literature (b) inconsistent comparisons being made in prior work (c) prior work showing performance brittleness across different goals and datasets. We argued that this creates an issue for businesses looking to use a microservice partitioning tool for internal software. Through Monte Carlo-style sampling, we chose a reduced, less correlated set of metrics, and used those as data for choosing weights for each goal, accounting for their different scales. We built upon an existing tool, CO-GCN (\cite{desai2021graph}), to build {\IT} in this paper, which fixes the issues listed above. To the best of our knowledge, ours is the first structured attempt at (a) reviewing the literature for a list of state-of-the-art microservice partitioning approaches (b) comparing all of them on the same datasets and the same metrics (c) fixing the performance brittleness issue using weighted losses and tuning. Finally, we discussed the broader impacts of the approach specified in this paper, generalizing the concept beyond the specific case of microservice partitioning.

Our approach is extensible and modular, consistently outperforms other approaches across datasets and metrics, and can easily be adapted to any metric that an enterprise is interested in. Moreover, the two loss functions (for the deep learner and the {\HPO} algorithm) can be tweaked to suit business goals. 

\section{Future Work}
\label{sec:future}

Our approach being modular leads to several avenues of future work, which we discuss in this section.

Because we apply weighted losses at the {\HPO} level, we can apply the same approach using a different base algorithm than CO-GCN. Specifically, we could build a Pareto frontier using a different state-of-the-art algorithm and then use our weighted loss function to choose a ``best" candidate.

Further, it would be useful to explore our methods on more datasets and metrics. In particular, it would be beneficial to test our methods on large enterprise systems. In addition, businesses might be interested in how much faster our system is compared to human effort.

Finally, our method
(using weighted losses to guide
exploration of the Pareto frontiers) is a general method that is not specific to microservice partitioning. Specifically, since our losses choose from a Pareto frontier generated by a hyper-parameter optimizer, (a) the choice of optimizer is left to the user (b) the application that the optimizer is applied to can be freely changed. Therefore our methods might offer much added value to other areas where {\HPO} has   been applied.

\section*{Acknowledgments}

This research was partially  funded by  
an IBM Faculty Award.
The funding source had no influence on the study design, collection and analysis of the data, or in the writing of this report.

\bibliographystyle{model5-names}\biboptions{authoryear}
\bibliography{cite}

\end{document}